\documentclass{article}
\usepackage{spconf,cite}
\usepackage{amsmath,amssymb}
\usepackage{graphicx}
\usepackage{booktabs}
\usepackage[hidelinks]{hyperref}

\begin{document}
\ninept

\title{FROZEN FLOWS FORGET: DIAGNOSING AND RESTORING LOST MOTION IN A LATENT-FLOW WORLD MODEL}

\name{Xiwen Chen$^{1,2,*}$\quad Rigaudiere Li$^{1,*,\dagger}$\quad Zhiruo Zhou$^{1,3,*}$\quad Xiaojun Zhu$^{1}$\quad Houde Liu$^{1,\dagger}$}
\address{$^{1}$Shenzhen International Graduate School, Tsinghua University\\
$^{2}$Shanghai Jiaotong University\\
$^{3}$Wuhan University of Technology\\
$^{*}$Equal contribution.\quad$^{\dagger}$Corresponding author.}

\maketitle

\begin{abstract}
Latent world models that integrate a flow in a frozen self-supervised latent space train stably and cheaply, yet silently lose the property manipulation depends on most: motion. The pretrained flow never moves the manipulated object; retraining it with latent-only losses only trades stillness for teleport-like motion. We trace the failure to the training signal, not the representation: anchor-sparse, latent-only supervision never says where along the horizon change belongs. Decode-augmented rollout training (DART) repairs this while keeping the representation frozen, retraining only the flow with decode-path supervision. DART outperforms its latent-only parent on the full protocol, restores the temporal structure of motion, and re-couples predicted motion to the scene; at larger scale it further improves prediction quality, closing nearly half the remaining gap to an oracle-informed interpolation reference. Finally, we report an unexpected finding about evaluation: pixel error alone rewards frozen predictions.
\end{abstract}


\section{Introduction}
\label{sec:intro}

Latent world models predict future observations by learning dynamics in a compact representation space instead of pixel space~\cite{dreamer,dreamerv3,diamond}.\footnote{Code: \url{https://github.com/Mangkhut160/icassp2027}. Trained checkpoint: \url{https://huggingface.co/pzr114514/DART-PT-Flow-LIBERO}.} One influential design freezes a self-supervised encoder entirely and learns only a flow that integrates the latent over time; ODEWorld~\cite{odeworld} instantiates this on LIBERO~\cite{libero} with a DINOv2~\cite{dinov2} reconstruction autoencoder (RAE). Freezing stabilizes training and shrinks the trainable surface---but removes every pathway through which training can be forced to preserve what matters.

\begin{figure}[t]
\centering
\includegraphics[width=\linewidth]{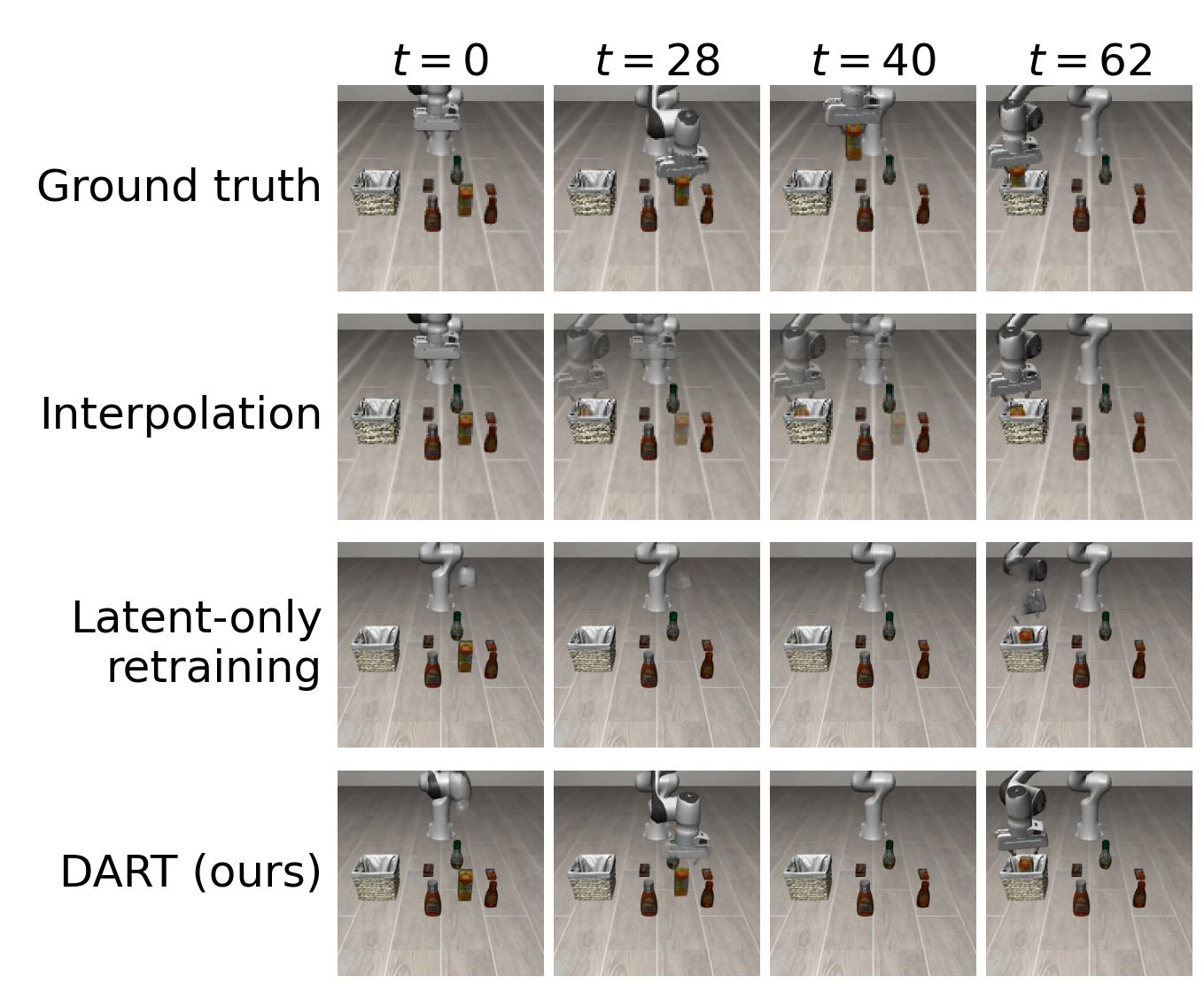}
\caption{64-step open-loop rollouts on the same demonstration. Linear interpolation between the endpoints ghosts the moving object and never resolves the grasp; latent-only retraining loses the grasped object during transport (absent at $t{=}28$--$40$) and it reappears only in the basket; DART keeps the object in frame across the whole rollout.}
\label{fig:teaser}
\end{figure}

On this system we document a silent degradation, invisible to standard losses yet obvious to any observer: \emph{motion collapse} (Fig.~\ref{fig:teaser}). Open-loop rollouts leave the scene essentially static, teleport the object in a few steps, and lose the manipulator arm for most of the horizon---the model has, in effect, unlearned that scenes move at all.

\begin{figure*}[t]
\centering
\includegraphics[width=0.86\textwidth]{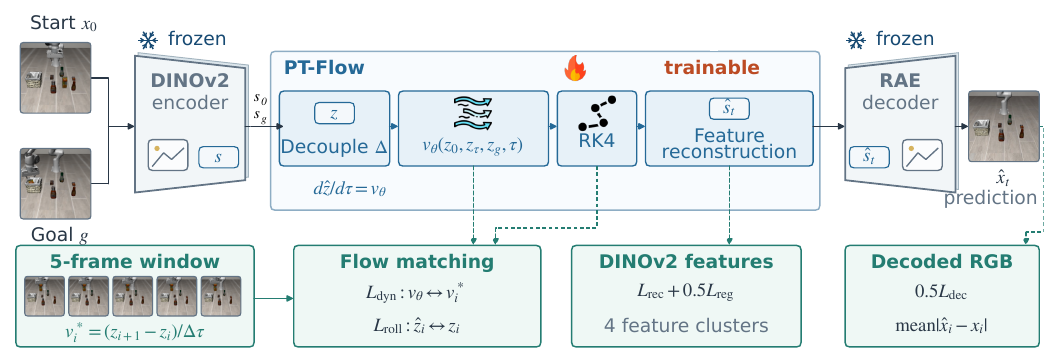}
\caption{Image-goal DART: frozen DINOv2 and RAE surround trainable PT-Flow modules; dashed paths connect predictions to losses. The window is illustrative and uses a validation demonstration that did not take part in training.}
\label{fig:method}
\end{figure*}

We diagnose the failure to its root cause and repair it with the smallest possible intervention---the representation stays frozen throughout (Fig.~\ref{fig:method}). We first rule out the natural suspect, a train/inference time-scale mismatch: conditioning rollouts on more intermediate goals degrades monotonically. The actual cause is that supervision is both anchor-sparse and latent-only: about five anchors per 275-frame demonstration, all graded in latent space alone, leave the flow free to park motion anywhere along the horizon. Our repair, \emph{decode-augmented rollout training} (DART), keeps the sampling schedule and instead changes what the supervision is graded in, extending the objective to decoded frames and region-level DINOv2 features so the gradient reaches the flow in the space the viewer sees. We then place the repaired system on an absolute scale, and show that the most natural evaluation metric---pixel error---systematically rewards the failure we set out to fix.

Our contributions:
\begin{itemize}\itemsep1pt
\item A quantified diagnosis of motion collapse in frozen latent-flow designs: open-loop rollouts freeze the scene and teleport the object, with teleport statistics, template tracking, and a motion-concentration metric localizing the cause to anchor-sparse, latent-only supervision.
\item DART, which retrains only the flow with decode-path supervision, outperforming its latent-only parent on the full protocol, restoring motion structure with one fixed weighting across six runs, and re-coupling predicted motion to the scene (Sec.~\ref{sec:exp}).
\item External reference points that fix the system's absolute level---the original paper's reported external world models, an oracle-informed interpolation floor, a trained video predictor in our harness, and the frozen-decoder upper bound---at which retraining closes $37\%$ of the interpolation gap and DART $47\%$ (Sec.~\ref{sec:anchors}).
\item An unexpected finding about how these models are evaluated: pixel L1 \emph{rewards stillness}. The pixel loss is minimized by the near-static conditional mean of an uncertain future, so a flow that moves less scores better while erasing the motion it exists to predict---the variant in our family with the \emph{lowest} L1 is also the most static one, and would have been selected as best by L1 alone.
\end{itemize}

\section{Related Work}
\label{sec:related}

World models differ in where their dynamics live: latent imagination drives control in Dreamer and DreamerV3~\cite{dreamer,dreamerv3}, DIAMOND~\cite{diamond} shows how much pixel-space detail matters in Atari, and generative world models at scale~\cite{genie,gamengen} renew the architecture debate; video metrics such as FVD~\cite{fvd} and multi-dimensional suites~\cite{vbench} grade the predictions such models produce. A long-standing pathology shadows all of them: pixel-trained predictors regress to the blurry conditional mean~\cite{mathieu,sv2p,savp}, a bias we show reappears at the level of rollout \emph{dynamics}. On LIBERO specifically, recent video world models~\cite{diff2026} find pixel metrics and action relevance largely orthogonal---independent support for our claim that pixel error cannot be the sole referee. We take the ODEWorld system as given and complement architecture work with failure analysis, a minimal repair, and motion-aware evaluation.

\section{Diagnosing Motion Collapse}
\label{sec:diag}

\subsection{Preliminaries}
\label{sec:prelim}
We first restate the pipeline that the diagnosis below applies to. A frame $x_t$ is encoded by a frozen RAE encoder $e$ (DINOv2) to a latent $z_t$; a flow $f_\theta$ integrates $\mathrm{d}z/\mathrm{d}\tau=f_\theta(z,\hat{g})$ over flow time $\tau\in[0,1]$---flow time measures how far the latent has traveled toward a goal latent $\hat{g}$ (an image goal, or one predicted from language); the frozen decoder $d$ renders frames, and an open-loop rollout decodes $T$ such increments from the encoded start frame without feedback. The key asymmetry for what follows is that demonstrations span up to 275 frames, but training anchors supervision at only about five latent waypoints per demonstration.

\subsection{Motion collapse}
\label{sec:motiondiag}
\begin{figure}[t]
\centering
\includegraphics[width=0.82\linewidth]{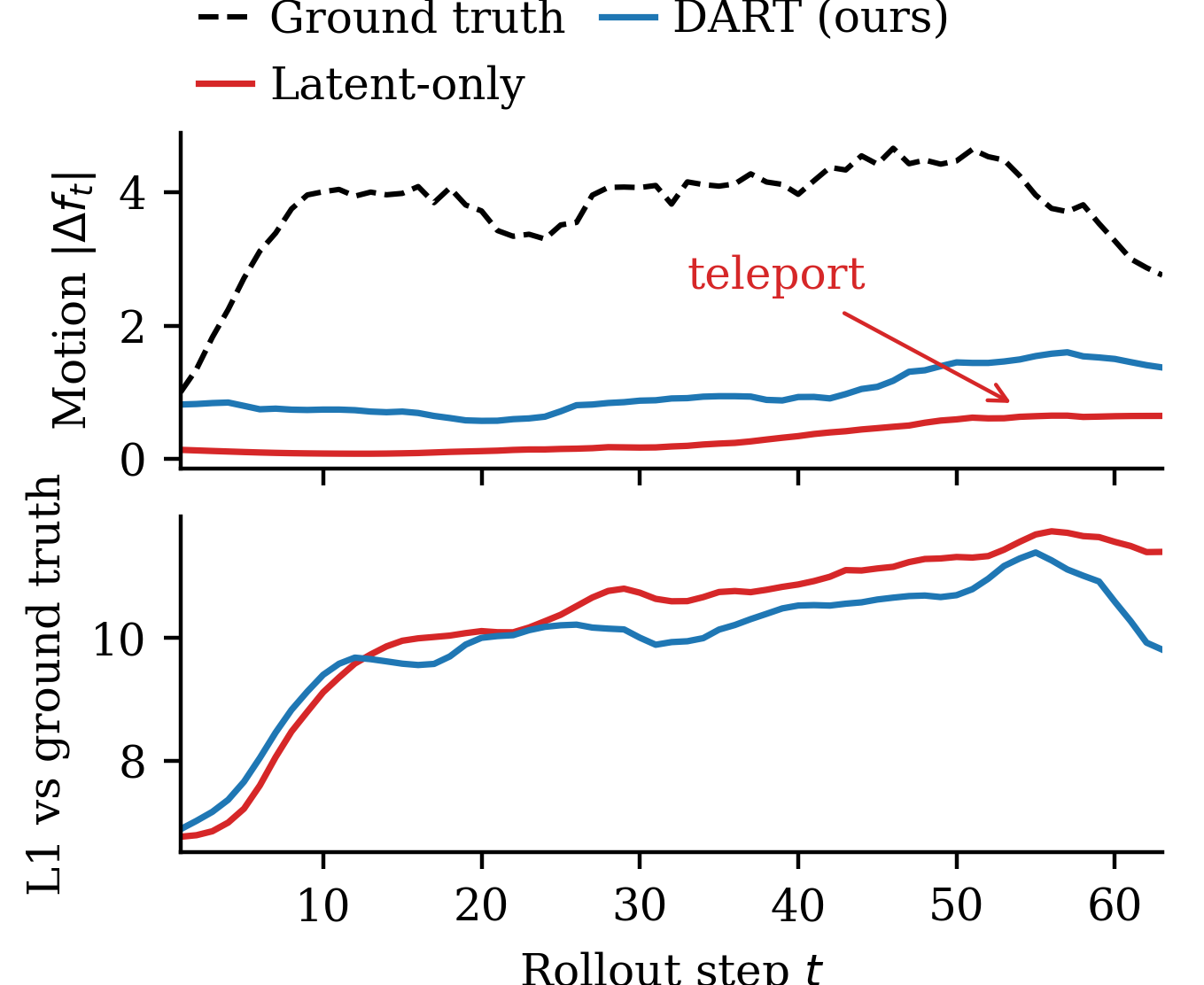}
\caption{Per-step motion (top) and error against ground truth (bottom), averaged over the 100-demonstration protocol. Latent-only retraining is flat and then spikes (teleport); DART tracks the ground-truth motion profile and stays closer in L1.}
\label{fig:diagnosis}
\end{figure}

This asymmetry shows up directly in the rollouts. Template tracking of the target object across open-loop rollouts (Fig.~\ref{fig:diagnosis}) separates three behaviors. Ground truth moves the object smoothly. The pretrained flow \emph{never moves the object at all}. Retraining that flow with latent-only losses moves it but teleports: $68\%$ of tracked displacement lands in three steps, the manipulator stays absent until the $\tau{=}1$ goal conditioning takes over, and a single step carries the object $36$\,px.

We quantify the shape of motion with two complementary statistics, both reported throughout. The \emph{tracked-displacement concentration} is the share of the object's displacement inside its three largest steps---intuitive, but needing a per-object tracker. The \emph{frame-level motion concentration} $\mathrm{conc}_3=\big(\sum_{t\in\mathcal{T}_3}\bar{\Delta}_t\big)/\sum_t \bar{\Delta}_t$ is the share of total frame-to-frame change $\bar{\Delta}_t=\mathrm{mean}|x_{t+1}-x_t|$ captured by the three largest steps; this one needs no tracker and is computable on every frame.

Before adopting the sparsity explanation, we rule out the natural suspect, a train/inference time-scale mismatch: conditioning rollouts on $S$ evenly spaced intermediate goals should help if scale were the cause, but error instead grows monotonically with $S$ (L1 $11.08{\to}47.60$ for $S{=}1{\to}64$; 41 demonstrations), because near-goal conditioning is out of distribution for a flow trained on whole-demo spans. The cause is anchor sparsity combined with latent-only losses: nothing tells the flow \emph{where along the horizon} change must happen.

\section{DART: Decode-Augmented Rollout Training}
\label{sec:method}

DART leaves the sampling schedule unchanged: the anchors keep the same positions and spacing, and only the space in which the supervision is applied changes. The loss adds two decode-pathway terms to the latent rollout objective,
\begin{equation}
\mathcal{L}=\mathcal{L}_{\mathrm{lat}}+0.5\,\mathcal{L}_{\mathrm{dec}}+0.5\,\mathcal{L}_{\mathrm{reg}},
\label{eq:dart}
\end{equation}
where $\mathcal{L}_{\mathrm{lat}}$ is the latent rollout error over the window, $\mathcal{L}_{\mathrm{dec}}=\mathrm{mean}\,|d(\hat{z}_t)-x_t|$ supervises decoded frames, and $\mathcal{L}_{\mathrm{reg}}$ matches frozen DINOv2 features of predicted and reference frames, pooled over four per-frame feature clusters. Encoder, decoder, and goal predictor stay frozen; only the flow trains, with AdamW (learning rate $10^{-5}$, weight decay $10^{-4}$, batch size 2) for 26{,}000 steps. DART must therefore satisfy three criteria at once: lower full-protocol L1 than its parent, motion concentration at or below the target fixed before training, and ordinary rollouts no worse. The gain comes from where the loss lands. Both new terms decode before the error is computed, so a mismatch visible in the picture becomes a loss directly, and the gradient travels back to the flow along the decode path. The latent-only loss, by contrast, compares latent values alone and cannot tell whether the picture actually moved.

\section{Experiments}
\label{sec:exp}

\subsection{Setup}
All experiments are offline open-loop predictions on official LIBERO demonstrations: 64-step image-goal rollouts at $256{\times}256$. We use three evaluation protocols. The \emph{full} protocol scores 100 demonstrations, with DART and latent-only runs matched by training seed. The \emph{hard-5} subset collects the five demonstrations with the highest parent error, selected once before any DART training and shared by every variant; it also serves as our stress test, since the parent errs most where the ground truth moves most. A 650-demonstration benchmark tests scale. Motion is graded by three statistics: per-step L1 catches wrong pixels, $\mathrm{conc}_3$ catches mistimed motion, and total motion catches missing motion. We also report the per-demonstration correlation between predicted and ground-truth motion.

\begin{table}[t]
\centering
\caption{Full 100-demonstration protocol (GT $\mathrm{conc}_3{=}0.093$). Med.\ L1 is the median over all frame-level errors of the same evaluation; total motion is $\sum_t\bar{\Delta}_t$, higher is closer to ground truth.}
\label{tab:full}
\footnotesize
\setlength{\tabcolsep}{3pt}
\begin{tabular}{lcccc}
\toprule
Model & L1 $\downarrow$ & Med.\ L1 $\downarrow$ & $\mathrm{conc}_3$ $\downarrow$ & Motion $\uparrow$ \\
\midrule
Latent-only (seed A) & 10.311 & 9.64 & 0.345 & 26 \\
DART (seed A) & \textbf{9.944} & 9.30 & \textbf{0.145} & \textbf{69} \\
Latent-only (seed B) & 10.214 & 9.53 & 0.173 & 58 \\
DART (seed B) & 10.015 & \textbf{9.29} & 0.225 & 47 \\
\bottomrule
\end{tabular}
\end{table}

\begin{table}[t]
\centering
\caption{Hard-5 subset (GT $\mathrm{conc}_3{=}0.099$), identical five demonstrations for every variant; ranges span independent runs.}
\label{tab:hard5}
\footnotesize
\setlength{\tabcolsep}{3pt}
\begin{tabular}{lccc}
\toprule
Model & L1 $\downarrow$ & $\mathrm{conc}_3$ $\downarrow$ & Motion $\uparrow$ \\
\midrule
Latent-only parent & 14.03--14.21 & 0.236--0.447 & 21 / 58 \\
DART & \textbf{13.68--13.91} & \textbf{0.145--0.272} & \textbf{41--64} \\
DART, decode $w{=}1$ & 13.87 & 0.220 & 48.6 \\
DART, region $w{=}1$ & 13.91 & 0.218 & 45.5 \\
Part decomposition & \underline{13.52}--13.86 & 0.309--0.445 & 19--31 \\
~~+ part-flow loss & 13.72--13.79 & 0.394--0.400 & 24--26 \\
\bottomrule
\end{tabular}
\end{table}

\subsection{Does DART restore motion?}
On the full protocol (Table~\ref{tab:full}), DART improves on its latent-only parent in mean and median L1 on both seeds. The improvement in motion structure is clearest on the first seed pair: concentration falls to $0.145$ against ground truth $0.093$, total motion rises from $11\%$ to $29\%$ of ground truth, and the largest single-step change stays at ground-truth scale, so motion spreads across the horizon instead of forming new teleports. The second seed keeps the error gain with a smaller motion shift, within the hard-5 run-to-run spread. The two motion metrics capture different aspects of the rollout: DART largely fixes \emph{where} motion happens but not \emph{how much}. No term in the objective rewards the magnitude of change, so a persistent gap in magnitude is expected.

On the hard-5 subset (Table~\ref{tab:hard5}), all six DART runs improve on both parent runs in L1, and the family-mean concentration drops from $0.342$ to $0.228$, better than the stronger parent run in five of six runs. The runs share the same recipe and the same five demonstrations, so their spread comes only from optimization randomness. Doubling either loss weight barely changes the result (Table~\ref{tab:hard5}), so we keep both at $0.5$. The relation between motion and scene content is more telling. In the parent, per-demonstration total motion is uncorrelated with ground truth (Spearman $-0.07$), so how much it moves carries no information about what actually moves and is instead an artifact of the model; DART restores the relation (Pearson $+0.26$), and predicted motion is again driven by the scene rather than by the model itself.

\subsection{Does pixel L1 mislead?}
\begin{figure}[t]
\centering
\includegraphics[width=0.66\linewidth]{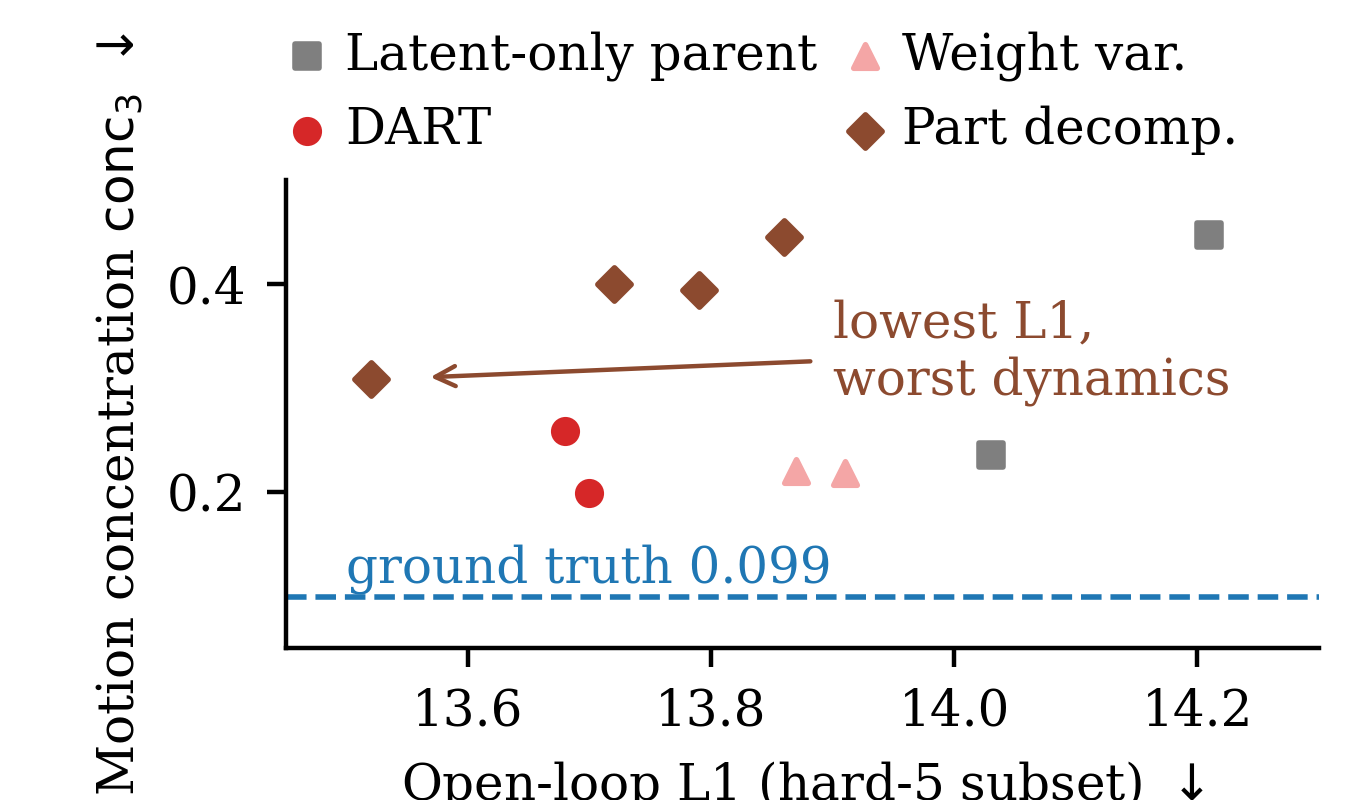}
\caption{The stillness trap (hard-5): part-decomposition variants attain the lowest L1 but revert to teleport-like concentration; DART reaches the best dynamics at comparable L1.}
\label{fig:trap}
\end{figure}

The part-decomposition variant gives the clearest warning. It replaces DART's region loss, which pools DINOv2 features over four clusters computed within each frame, with a decomposition whose parts keep the same identity across frames; it also widens the window to 9 and optionally adds a loss on the direction in which parts displace. It reaches the lowest L1 of any variant we trained and the worst dynamics in the family (Fig.~\ref{fig:trap}). The reason is that the future is uncertain: many next frames are plausible, and their average is blurry and nearly static, so L1 is minimized by predicting close to that average. A flow that moves less therefore scores better on L1 while destroying the motion it exists to predict. Pixel losses are known to induce this blur bias in video prediction~\cite{mathieu,sv2p}, and here it is visible across the entire rollout. Ranked by L1 alone, it would have been selected as the best in the family, which is why the concentration target was fixed before training.

\subsection{How does the system compare in absolute terms?}
\label{sec:anchors}
\begin{table}[t]
\centering
\caption{Absolute reference points. Top: external models as reported in ODEWorld's comparison (their protocol, 6{,}500 demos)~\cite{odeworld}. Middle: our harness, 650-demo split; interpolation is oracle-informed, RAE is a decode upper bound, not predictors. Bottom: SimVP~\cite{simvp}, same split (two-seed ranges); LPIPS is Alex, rollout-averaged.}
\label{tab:anchors}
\footnotesize
\setlength{\tabcolsep}{3pt}
\begin{tabular}{lcc}
\toprule
Reference & PSNR $\uparrow$ & LPIPS $\downarrow$ \\
\midrule
\multicolumn{3}{l}{\emph{Reported in~\cite{odeworld}, step 64, their protocol:}}\\
LDP (latent diffusion) & 16.27 & 0.461 \\
V-JEPA 2 (frozen-encoder) & 16.47 & 0.166 \\
\midrule
\multicolumn{3}{l}{\emph{Our harness, 650-demo split, mean over 64 steps:}}\\
Pretrained ODEWorld & 17.65 & 0.262 \\
Latent-only retraining (3 seeds) & 19.24--19.45 & 0.196--0.212 \\
DART (3 seeds) & 19.67--19.77 & 0.190--0.195 \\
Linear interpolation (oracle-informed) & 21.96 & 0.159 \\
RAE reconstruction (upper bound) & 26.66 & 0.115 \\
\midrule
\multicolumn{3}{l}{\emph{Trained baseline in our harness, same split:}}\\
SimVP, interpolated input & 23.38--23.40 & 0.120--0.121 \\
~~+ DINOv2 feature loss & 23.49--23.52 & 0.118--0.120 \\
SimVP, endpoints only & 23.11--23.19 & 0.121 \\
\bottomrule
\end{tabular}
\end{table}

All comparisons so far have been internal, between DART and its own parent. We now place the system on an absolute scale against four references: two external world models reported by the original ODEWorld evaluation~\cite{odeworld}, an oracle-informed interpolation and the frozen decoder's reconstruction bound from our harness, and a trained video predictor (Table~\ref{tab:anchors}). The design we build on already led its published comparison, both external models trailing it there, and our improvements come on top: within our harness (PSNR and LPIPS~\cite{lpips}), flow retraining closes $37\%$ of the distance between the pretrained flow ($17.7$\,dB) and an oracle-informed linear interpolation ($22.0$\,dB), and DART closes $47\%$, about two decibels above the pretrained flow; the frozen decoder's reconstruction bound ($26.7$\,dB) localizes the remaining headroom in the latents the flow produces.

The most informative reference is SimVP~\cite{simvp}, a standard convolutional video predictor trained on the same split in our harness: it reaches $23.4$\,dB under start--goal--interpolation conditioning, matching ground-truth motion timing ($\mathrm{conc}_3$ $0.095$ vs.\ $0.101$) and motion tracking the scene (Spearman $0.88$). This exceeds our system: the task is learnable to a pixel level we do not attain, so the deficit is in the training signal, not the task. Could this match be leaked through the interpolated conditioning? A variant receiving only the two endpoints, with the rest zeroed, loses less than $0.3$\,dB and still reproduces both motion metrics: the match is learned, not leaked.

\subsection{Does it hold at scale?}
\label{sec:scale}
\begin{figure}[t]
\centering
\includegraphics[width=0.84\linewidth]{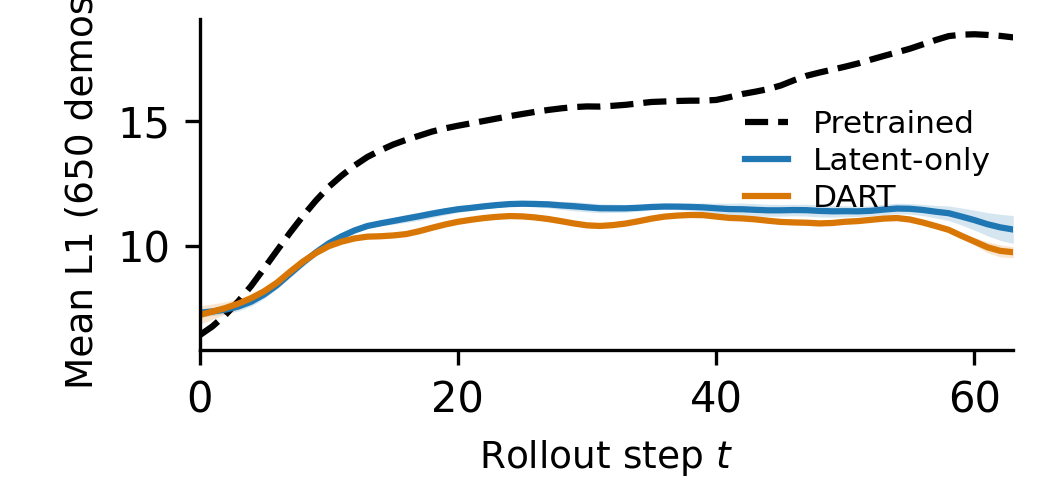}
\caption{650-demo, 64-step open-loop benchmark: per-step mean L1 for the pretrained, latent-only, and DART flows (3 seeds each, $\pm$1 std).}
\label{fig:benchmark}
\end{figure}

{\emergencystretch=1.5em
On 650 demonstrations (Fig.~\ref{fig:benchmark}), latent-only retraining of the same flow cuts per-step L1 by ${\approx}27\%$, and DART improves again at this scale: with three seeds per family, every DART seed falls below every parent seed (mean L1 $10.36$--$10.44$ against $10.73$--$11.01$). Gains are largest at the horizon's end, where the anchor-sparsity diagnosis predicts, and hold across the five LIBERO suites. Scaling sharpens pixels but leaves the magnitude gap, so we tested the remedy directly: two variants add a motion-magnitude term, on decoded consecutive-frame differences or on latent displacement norms, inside the DART recipe at matched budget. Neither works. The latent variant leaves full-protocol mean L1 unchanged ($9.88$ vs.\ $9.94$) while total motion falls to a third of DART's and concentration worsens; the decoded variant damages concentration at a quarter of the budget. Rewarding magnitude thus pulls the flow toward stiller rollouts: the gap reflects the objective's regression structure, not a missing loss term.
}

\subsection{Failure analysis and future work}
Beyond the magnitude gap, the weakest link is the intermediate frames: decoded rollout frames saturate near $19$--$20$\,dB PSNR late in the horizon while the same decoder renders ground-truth latents at about $27$\,dB, so the deficit is in the produced latents. LPIPS objectives, region-level losses, velocity-magnitude proxies, longer budgets, and wider windows each fail to gain beyond about a percent, and a motion-magnitude reward (Sec.~\ref{sec:scale}) lowers motion instead of raising it. Two causes are plausible: the regression to the mean behind the magnitude gap, and a bottleneck, since recognition-tuned DINOv2 latents may discard texture needed for frame sharpening. The matching remedies point to future work beyond the objective: blur-punishing decoders or encoder-side latent enrichment, plus motion-magnitude teachers such as optical-flow supervision.

\section{Conclusion}
Frozen latent-flow world models can pass every training loss while forgetting how scenes move. The failure is diagnosable (anchor-sparse, latent-only supervision) and repairable without unfreezing the representation, and because pixel error rewards frozen predictions, motion-aware supplements are needed at evaluation.

\end{document}